**Title:** The Point of View of a Sentiment: Towards Clinician Bias Detection in Psychiatric Notes
**Authors:** Alissa A. Valentine[1], Lauren A. Lepow[2], Lili Chan[1], Alexander W. Charney[1], Isotta Landi[1]
**Affiliations:** [1]Icahn School of Medicine at Mount Sinai, Institute for Personalized Medicine; [2]Icahn School of Medicine at Mount Sinai, Department of Psychiatry

**Abstract**

Negative patient descriptions and stigmatizing language can contribute to generating healthcare disparities in two ways: (1) read by patients, they can harm their trust and engagement with the medical center; (2) read by physicians, they may negatively influence their perspective of a future patient. In psychiatry, the patient-clinician therapeutic alliance is a major determinant of clinical outcomes. Therefore, language usage in psychiatric clinical notes may not only create healthcare disparities, but also perpetuate them. Recent advances in natural language processing systems have facilitated the efforts to detect discriminatory language in healthcare. However, such attempts have only focused on the perspectives of the medical center and its physicians. Considering both physicians and non-physicians' point of view is a more translatable approach to identifying potentially harmful language in clinical notes.

By leveraging pretrained and large language models (PLMs and LLMs), this work aims to characterize potentially harmful language usage in psychiatric notes by identifying the sentiment expressed in sentences describing patients based on the reader's point of view. Extracting 39 sentences from the Mount Sinai Health System containing psychiatric lexicon, we fine-tuned three PLMs (RoBERTa, GatorTron, and GatorTron + Task Adaptation) and implemented zero-shot/few-shot in-context learning (ICL) approaches for three LLMs (GPT-3.5, Llama-3.1, and Mistral) to classify the sentiment of the sentences according to the physician or non-physician point of view.

Results showed that GPT-3.5 aligned best to physician point of view and Mistral aligned best to non-physician point of view, both with an ICL approach. These results underline the importance of recognizing the reader's point of view, not only for improving the note writing process, but also for the quantification, identification, and reduction of bias in computational systems for downstream analyses.

**Introduction**

Psychiatric notes document the clinical signs, symptoms, and behaviors of patients from the perspective of the clinician. When documenting the clinical encounter, the language used by clinicians can be classified as neutral, negative, or positive[1]. Negative patient descriptors include those that question patient credibility, reasoning, insight, or judgement; portray the patient as noncompliant or as a threat; remark on the patient's poor self-care; or generally conveys disapproving feelings towards the patient and their presentation. In contrast, positive patient descriptors include patient strengths, minimization of blame, and language that conveys of approval and positive feelings towards the patient and their presentation.

The use of negative language in clinical notes carries two distinct downstream harms which may sway patient outcomes. When read by future providers, harmful language use can impact their perspective of a patient and decrease their quality of care[2]. When read by patients, inaccurate or negative patient descriptions fosters mistrust and harms the therapeutic alliance[3-5], a major determination of positive outcomes in psychiatry[6-8]. Therefore, in both scenarios, while not necessarily intended by the writer, there is potential for harm by perpetuating biases from the medical system, in a way that hinders mutual engagement and connection between patients and their clinicians.

With the ubiquitous use of natural language processing (NLP) systems in healthcare research[9], the use of harmful language in clinical notes also threatens equitable deployment of artificial intelligence (AI) in medical contexts. Recent work has shown that societal biases are often reflected in AI. Namely, if biased language is embedded in the corpora used to train large language models (LLMs), models learn to perpetuate societal biases across gender, language, race, ethnicity, and insurance status[10,11]. Without taking appropriate action, these models risk contributing to perpetuating health disparities, as seen in the statistically different performance rates on clinical prediction tasks between sociodemographic groups[12,13].

Current definitions of harmful language include words from discriminatory and stigmatizing lexicons[1,2,14,15]. Attempts to assess harmful language in health care settings have recently increased, leveraging such lexicons and benefiting from the advances in NLP systems[16-18]. These approaches and lexicons rely on a consensus perspective, which has yet to be dissected into how clinicians and non-clinicians perceive the same words and their usage. One explanation could be that taking the patient's perspective into account is a newer issue. In 2020, the 21st Century Cures Act went into effect, granting patients the right to immediately access their electronic health record data, including physicians' notes, during a clinical encounter[19]. Patient-facing interfaces, such as MyChart, furthermore make clinical notes easily accessible for reading—some even alert the user that a new note is ready to view. The perspectives of patients must therefore be considered when flagging potentially harmful language, if we aim to holistically address its downstream effects.

Sentiment analysis in NLP is the process of determining whether the tone conveyed by the written text is positive, neutral, or negative, and it has been leveraged to identify harmful language such as hate speech[20]. As such, sentiment analysis can be used as a proxy in the clinical domain to identify discriminatory and stigmatizing language use. Yet, existing sentiment analysis methods have not been optimized for use in the clinical domain, particularly in psychiatry. Recent approaches to implementing sentiment analysis rely on sentiment lexicons of negative and positive words or sentences to label publicly available clinical note data[21,22]. These existing methods exhibit low validity and high variability and do not generalize to psychiatry[23-25]. Furthermore, there is a lack of psychiatric clinical note datasets with sentiment annotation, and no existing annotations include the perspectives of patients. Finally, existing sentiment analysis datasets consist of single consensus labels amongst annotators, wherein there is only one correct sentiment per data point. This limits the ability to explore subjectivity and reflect real world scenarios when more than one label is correct to describe a patient[26].

To this end, we investigated how language models capture the subjective point of view of physicians and non-physicians towards sentences containing patient descriptions from psychiatric clinical notes. Using a sentiment analysis task, we explored which language models performed best at classifying sentences as negative, neutral, or positive from the physician or non-physician point of view. We built upon a real-world lexicon of patient descriptions used in naturalistic settings[15] to aid our work, tailoring it for the psychiatric setting. We evaluated the performance of three LLMs (i.e., GPT-3.5, Llama-3.1, and Mistral) on the same sentiment analysis task via prompt-based approaches and in-context learning (ICL). As a baseline, we fine-tuned three pretrained language models (PLMs), i.e., RoBERTa, GatorTron, and GatorTron, with task adaptation (TA), on a sentiment analysis classification task.

This work demonstrates that the successful deployment of LLM sentiment analysis methods in healthcare depends on addressing the differing points of view between physicians and patients and the subjectivity of sentiment labels that reflect real-word scenarios in psychiatry. In doing so, we learn how to optimize LLMs as-is for tasks using psychiatric text and move towards understanding how clinician bias is represented in the clinical text input into LLMs. This work not only aims at informing the clinician note writing process, but also at providing insight into bias quantification, identification, and removal.

**Results**
We investigated whether physician's and non-physician's points of view could be captured by LLMs via sentiment analysis by asking physicians (n=10) and non-physicians (n=10) to label 39 rule-based extracted real-world sentences describing psychiatric patients. As a baseline comparison, we fine-tuned PLMs on the same task to investigate how PLMs pre-trained on clinical text (i.e., GatorTron) or fine-tuned on sentiment analysis task (i.e., RoBERTa) perform compared to non-specialized LLMs reinforced with human feedback to limit the generation of harmful content.

We also explored how subjectivity impacts model performance on the sentiment analysis task by evaluating the models on four datasets with different agreement levels: (1) the *Baseline* dataset including all sentences (n=39); (2) a *≥70% agreement* dataset including sentences with the corresponding rate of agreement within physician/non-physician annotations (n=33); (3) a *≥80% agreement* dataset (n=23); and (4) a *≥90% agreement* dataset (n=14). A fifth, *no agreement* dataset, was created to include sentences where the physician and non-physician label disagreed (i.e., the physician label is negative, and the non-physician label is positive; n=8). To determine agreement level within groups, a physician and non-physician label was assigned to each sentence based on majority voting. Each dataset, except for the *no agreement* dataset, was split into training and validation. The training sentences were used to fine-tune the PLMs and provide in-context learning to the LLMs. The validation sentences were used to evaluate the PLMs and LLMs. An external test set of new sentences describing psychiatric patients (n=15) was used to manually validate the PLMs and LLMs. See the Methods section for more details.

*Rater Agreement*
Our results demonstrate a nuanced subjectivity in how physicians and non-physicians perceive the sentiment of the same sentence such that non-physician labels were more polarized than physicians (**Figure 5**). Namely, physicians labeled more sentences as neutral compared to non-physicians (**Figure 1**; 54% vs 44% in the *Baseline* dataset) and had higher average agreement on neutral labels than non-physicians (**Error! Reference source not found.**; 69% vs 63%). Non-physicians demonstrated higher agreement on negative labels than physicians (**Error! Reference source not found.**; 79% vs 75%) and lower agreement on sentences containing clinical words, such as "compliant", "adherent", and "malingering" (**Figure 2**; 64.6% vs 71.4%). However, average percent agreement was lowest on neutral sentences for both physician and non-physician raters (66%). This resulted in a drop of neutral sentences included in the 80% agreement datasets, compared to the baseline and the 70% agreement datasets (**Figure 1**).

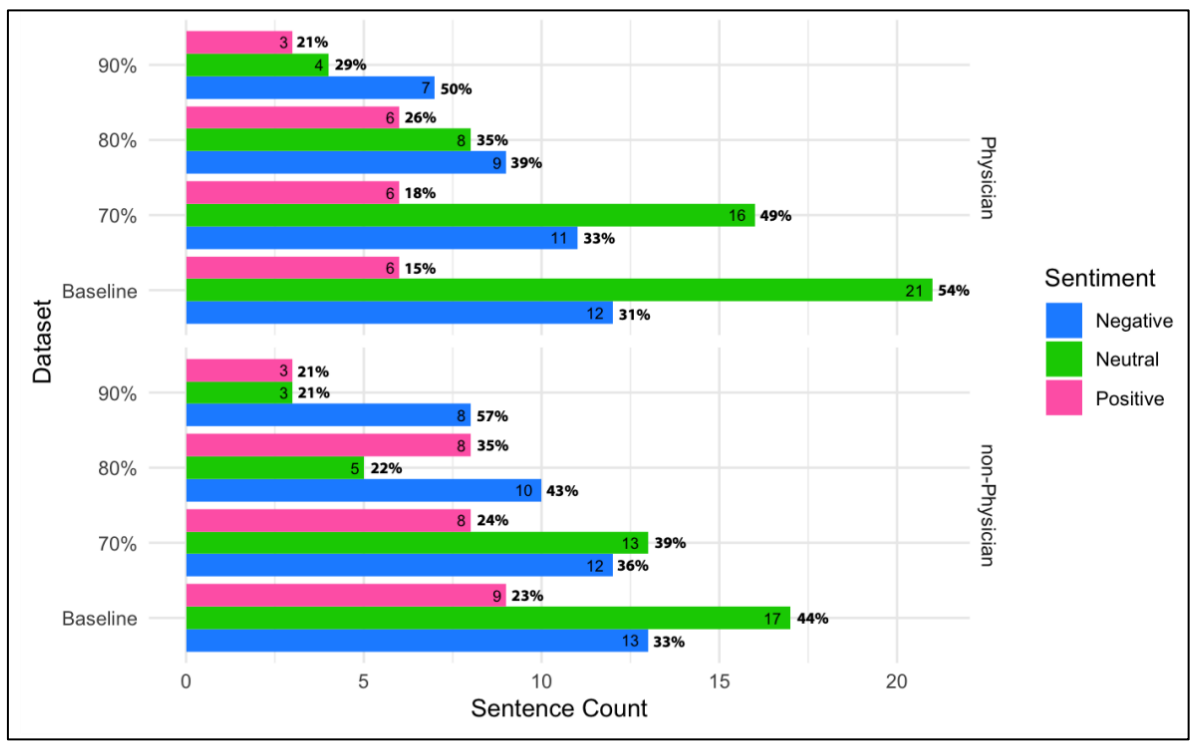

**Figure 1.** Counts and percents of negative, neutral, and positive sentences within each dataset for physician and non-physician labels. Datasets: Baseline, 70%: 70% Agreement, 80%: 80% Agreement, and 90%: 90% Agreement.

When examining label agreement between all 20 raters, we found fair agreement according to Fleiss classification[27], k = 0.323 (p < 0.001). A fair level of agreement was also found within physician raters (k = 0.321, p < 0.001) and non-physician raters (k = 0.345, p < 0.001). Good agreement was found when comparing the unified physician and non-physician labels with Cohen's Kappa coefficient (k = 0.673, p < 0.001), such that the unified labels were the same between the two groups 80% of the time.

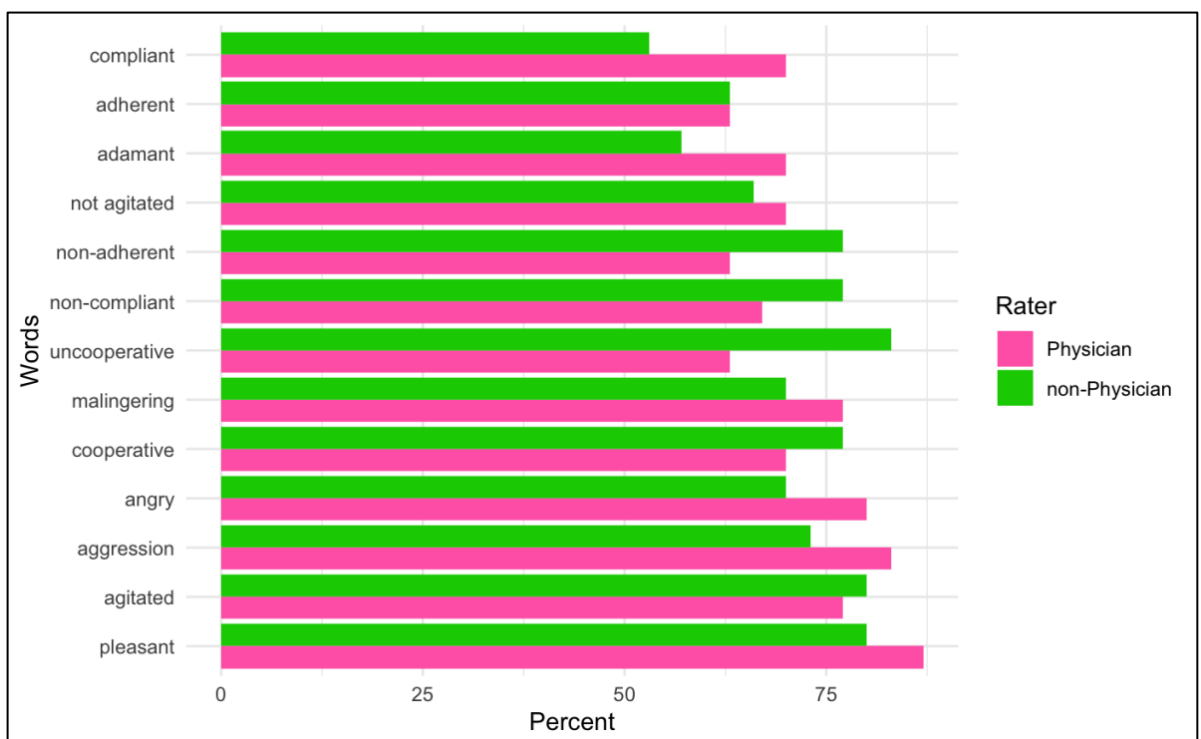

**Figure 2.** Word-level agreement percentage. Each bar denotes the percentage of the raters (physician or non-physician) that agreed with the unified label (negative, neutral, or positive) for the sentences containing each word in the lexicon. The words are in order of lowest to highest average agreement of both physician and non-physician labels, e.g., "compliant" carries the lowest average agreement across both physicians and non-physicians.

### *Language Models Performance*
The PLMs were fine-tuned to label the sentiment of sentences for the physician and non-physician labels of each Baseline, 70%, and 80% training datasets. PLMs were evaluated on the validation sentences from the corresponding dataset, while completing hyperparameter optimization. The external test set of new sentences was used to evaluate the fine-tuned PLMs.

The LLMs labeled the sentiment of sentences in the validation datasets using both a zero-shot and ICL approach sampling from the training sentences for the physician and non-physician labels of each Baseline, 70%, 80%, and 90% datasets. The test set evaluated each LLM with a zero-shot approach and the best performing ICL prompts.

The models' performance on the sentiment analysis task was measured via macro F1, precision, and recall scores on the validation and test sets. Mistral performed best on the non-physician labels (F1=0.93) and GPT-3.5 on the physician labels (F1=0.93).

*PLM Performance*
GatorTron and its task adapted version[28] (GatorTron-TA) outperformed RoBERTa on the validation and test datasets (**Table 2**). Task adaptation involved pretraining GatorTron on a small subset of leftover psychiatric sentences, see methods for more details.

During validation, the models performed better on the physician task (F1=0.90) than the non-physician task (F1=0.80). As seen in **Figure 6**, the validation performance of GatorTron-TA and RoBERTa improved on the non-physician task as agreement increased in the datasets. However, the opposite is seen in the physician task, such that as agreement increases the models' F1 score generally decreases during validation. When evaluating the models on the test set, they also performed better on the physician task (F1=0.77) rather than the non-physician task (F=0.71). During testing, GatorTron-TA performed best on the physician task when fine-tuned on the baseline sentences. GatorTron-TA and GatorTron tied for best performance on the non-physician task, but GatorTron-TA performed best when fine-tuned on sentences with ≥80% agreement and GatorTron performed best on the non-physician task when fine-tuned on Baseline sentences. As seen in **Figure 7**, the models had higher precision than recall on average. This demonstrates that the models had more false negatives than false positives, wherein negative and positive sentences were more likely to be mislabeled as neutral.

*LLM Performance*
Mistral and Llama-3.1 generally outperformed GPT-3.5 on both the physician and non-physician tasks with a zero-shot or ICL approach during validation. However, during evaluation on the test set, GPT-3.5 performed best on the physician task (F1=0.93) when using ICL with sentences with ≥80% agreement and Mistral performed best on the non-physician task (F1=0.93) when using ICL with sentences from the Baseline dataset.

With a zero-shot approach, the best performance was seen when evaluating Mistral on the non-physician labels of the validation sentences with ≥80% agreement (F1=0.94). The best performance on the physician labels was obtained by Llama-3.1 on the validation sentences with ≥90% agreement (F1=0.89). The results of the zero-shot approach demonstrate that the models' F1 scores increased on the validation datasets as the percentage of agreement on all the sentences increased (**Figure 8**) When evaluating the models on the sentences with no agreement between physicians and non-physicians, they all perform best on the non-physician labels. Interestingly, all three models demonstrate high precision (>0.80) when evaluated on the physician labeled no-agreement dataset with a zero-shot approach, but poor recall (<0.40). This implies that the sentences that only physicians perceive as negative or positive are often mislabeled as neutral.

When evaluating the models on the validation set with ICL, all three models reported an F1 score of 1.0 on the non-physician labels, whereas Mistral and Llama-3.1 reported an F1 score of 1.0 on the physician labels. The results of the ICL approach show that the models performed best on the validation sentences with ≥80% agreement (F1=1.0, **Figure 9**). Interestingly, all the models dropped in performance on the 70% validation dataset sentences with physician labels (**Table 4**). This decrease is driven by a large drop in recall, demonstrating that the models were mislabeling negative and positive sentences as neutral from the physician's point of view, similarly to the zero-shot results. The drop in F1 score is also due to an unbalanced number sentences per sentiment label, which is seen throughout the results, as the F1 score is lower than precision and recall for the physician labels in the 70% validation dataset.

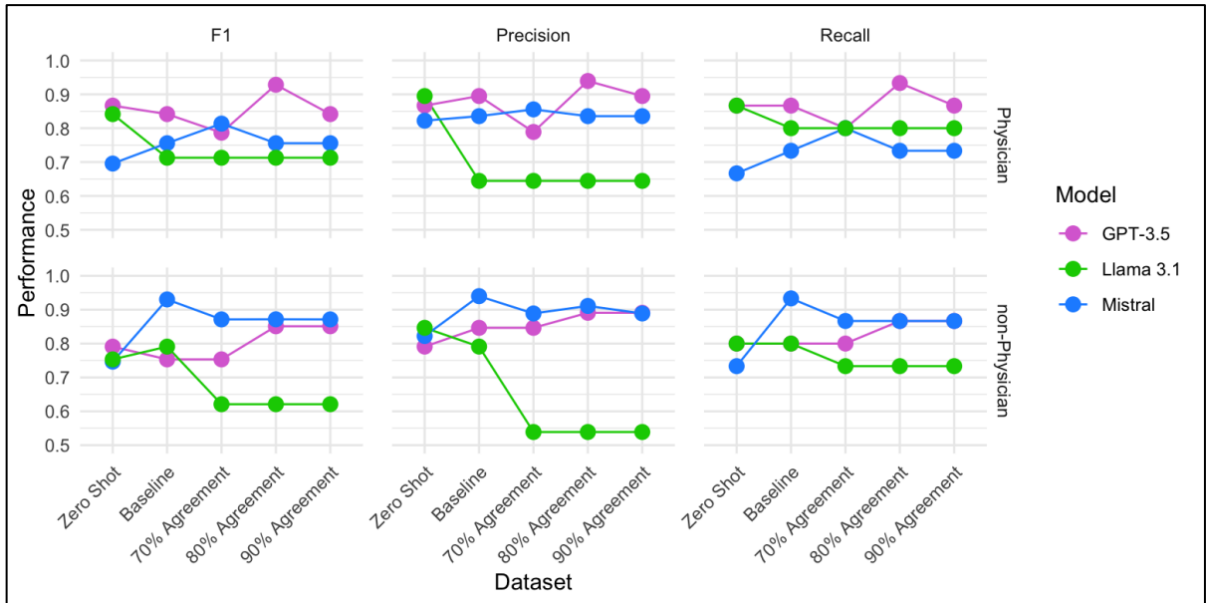

**Figure 3.** Large language model (LLM) performance on the test set (N=15 sentences). Each graph shows the performance of each LLM (i.e., GPT-3.5, Llama-3.1, and Mistral) on the test data with a zero-shot approach, or the best prompt corresponding to each dataset (i.e., Baseline, 70% Agreement, 80% Agreement, 90% Agreement). Macro F1 score, precision, and recall are reported.

During testing, ICL improved the performance of Mistral and GPT-3.5, but Llama-3.1 performed best with a zero-shot approach (**Figure 3**). This can be explained by the results of the prompt engineering task. **Figure 4** demonstrates the number of neutral, negative, and positive sentences that led to the best ICL prompt for each model and validation dataset during the validation task. Mistral and GPT-3.5 perform better when using more ICL sentences in their prompt such that both models typically used at least one ICL sentence. In contrast, Llama-3.1 doesn't perform better with ICL sentences apart from two scenarios, namely the baseline non-physician and 80% physician validation sentences. Therefore, when evaluated on the test set, Llama-3.1's ICL prompts only included example sentences when using the prompts with sentences from the non-physician baseline and physician ≥ 80% agreement training datasets. In every other scenario, the model is given a prompt that is structured to contain example sentences, but they are missing. Therefore, it is likely that this leads to a drop in performance when compared to the zero-shot approach. As seen in **Figure 3**, this prompt composition impacts the model's precision greater than its recall, thus increasing the model's false positive rate, i.e., when neutral sentences are mislabeled as negative or positive.

**Discussion**
In these experiments, we demonstrated that: (1) there are nuances between how physicians and non-physicians perceive the sentiment of sentences describing psychiatric patients in clinical notes; (2) LMs are sensitive to these nuances and show differential performance on physician and non-physician sentiment analysis tasks via token classification or prompt-based approaches; (3) LLMs outperform PLMs on sentiment analysis tasks with clinical text by 26%. When evaluating model performance on the test set, GPT-3.5 best classified the sentiment corresponding to the physician's point of view, and Mistral best classified the non-physician's point of view.

Our results suggest that there are differences in how physicians and non-physicians label the sentiment of sentences that go undetected by agreement statistics. Namely, physicians perceive more sentences as neutral than non-physicians, whose labels have more variation and more likely to be negative or positive (**Figure 1** and **Figure 5**). These differences could be explained by the training physicians receive, where they learn established clinical language and the medicolegal approach to documenting in the clinical notes, all influencing their intuitive clinical interpretation of the language. Nevertheless, this work's findings suggest that neutrality is the hardest sentiment for humans to agree on, and consequently for language models to label. We found that increased agreement is associated with fewer neutral sentences within our annotated datasets. Moreover, PLMs and LLMs struggled with labeling neutrality. GatorTron, with and without TA, GPT-3.5, and Mistral were "too neutral", i.e., negative and positive sentences were often mislabeled as neutral. Conversely, Llama-3.1 was more polarized, wherein neutral sentences were more likely to be mislabeled as negative or positive. These findings underline the importance of carefully investigating the sentiment composition in annotated text and determining whether to prioritize reducing false positives or false negatives in LLM outputs, particularly when deploying models for harmful language detection.

Furthermore, we demonstrated how different LMs align better to physician or non-physician points of view when tasked with the classification of the sentiment of sentences describing psychiatric patients. In our experiments, LLMs outperformed PLMs on the sentiment task we designed with only few examples, showcasing the feasibility of leveraging generative LMs to flag harmful descriptions of psychiatric patients from the perspective of physicians and non-physicians. This brings us closer to mitigating the impacts of harmful clinical language on health outcomes in psychiatry.

Due to a lack of transparency about pre-training data, it is difficult to speculate why certain LLMs align better to physician or non-physician points of view. Since Mistral performed best on the non-physician task, one explanation could be that Mistral's pre-training corpus does not include clinical text. Since GPT-3.5 and Llama-3.1 perform better on the physician task, we could hypothesize that their pre-training data includes clinical text. In contrast, the PLMs in this experiment are fully transparent about their pre-training data. GatorTron is pre-trained on over 2 million clinical notes[29] and RoBERTa is pre-trained on over 124 million tweets and finetuned on a sentiment analysis task[30,31]. Thus, it is expected that GatorTron would perform best on the physician-labeled dataset, whereas RoBERTa shows a poorer performance on clinical text overall. By task adapting GatorTron (i.e., further training it on clinical text from our own health care system), we further improved the model's performance on the physician task. PLMs also require fewer computational resources because their parameter count is in the millions, whereas LLMs have billions of parameters. The performance of the PLMs is impressive given how small they are in comparison to the LLMs. We encourage researchers to also evaluate PLMs on their tasks, as PLMs are the more sustainable approach.

In conclusion, this work advocates for the deployment of sentiment analysis methods in psychiatry to reflect real-world scenarios. This can be achieved by recognizing the point of view and subjectivity inherent to describing and perceiving the descriptions of psychiatric patients. Our work is the first to demonstrate how to leverage LLMs for such tasks and our contributions include:
- The generation of a manually curated lexicon of words and annotated dataset of sentences describing patients in psychiatry.

- The investigation of the differences between physicians' and non-physicians' sentiment towards sentences including descriptors.
- The implementation of an NLP framework to assess the alignment of LMs to physicians and non-physicians point of views.

These contributions bring us closer to utilizing methods that take into account the downstream harms of clinical language, and understanding how clinician bias is represented in clinical text.

**Limitations and Future Work**

Future work will consider cultural differences in the perspectives of sentiment towards psychiatric patients, as well as implement community-based approaches for recruiting active psychiatric patients to aid in defining clinician bias and negative patient descriptions. Throughout this project, we attempted to involve as many perspectives as possible when tasking raters with labeling the sentiment of sentences. However, due to privacy reasons, we could not collect personal data from raters, including their mental health or sociodemographic factors. It is possible that some of the non-physician raters are living with psychiatric diagnoses, but we are unable to make that assumption or claim that our work describes the point of view of active psychiatric patients. We consider this to be a large limitation, as psychiatric patients would be reading their own notes containing descriptions using psychiatric lexicon.

The long-term goal of our work is to explore approaches to debiasing language models and clinical note datasets that consider the point of view of the physician and non-physician. Once biased and harmful text is detected in clinical notes, the next task is deciding what to do about it. One solution is to "neutralize" language use, wherein harmful words are either removed or replaced with words conveying a neutral point of view[32]. However, no consensus exists on whose point of view, the patient or physician, should be used in such practices. Furthermore, we must work towards debiasing approaches that consider the context of the bias due to the reality that removing or replacing a word is not always the most ethical approach[33].

**Methods**

*Data*

*Patient Descriptor Lexicon*

A lexicon of psychiatric patient descriptors was created by LAL, a psychiatrist with experience in writing clinical notes and AAV, a computational scientist with experience in fairness, justice, and ethics in machine learning. A set of words (n = 54) was initially obtained combining a list of "never words" (i.e., discriminatory words or phrases that should never be used to describe patients in Emergency Medicine) created by the Mount Sinai Institute for Health Equity Research and the 15 patient descriptors listed in a highly cited paper about patient descriptors[24]. Words were then filtered to include:

(1) Words used exclusively to describe patients (e.g., "claims" was not included because it is often used for insurance information).
(2) Words often used to describe patient behavior in psychiatric settings (e.g., "agitated" is more often used to describe patients in psychiatric settings versus other medical settings).
(3) Words with subjective meaning, as determined by the aforementioned members of the research team (e.g., "compliance" carries different meaning depending on someone's clinical experience).

After filtering, 13 words, including negations, were retained (see
**Table 1**).

*Sentences Selection*

Three sentences matching each word from the lexicon of descriptors were extracted from an initial query of 1,000 random clinical notes from the Mount Sinai Health System's electronic health record database. Notes were limited to Progress Reports from clinical encounters where the billing diagnosis was a psychiatric diagnosis. Psychiatric diagnoses were defined as F01-99 International Classification of Disease, Tenth Revision (ICD-10) codes. A total of 39 sentences of length > 30 characters was obtained and protected health information (PHI) was manually masked (see examples in
**Table 1**).

*Physician and Non-Physician Annotations*
Members of our research institute, not involved in the study, labeled the sentences identified in Section 3.2. Physician raters (n=10) have medical degrees and extensive experience writing clinical notes. Non-physician raters (n=10) have no clinical experience, nor medical degree, and received no training on clinical note writing. Annotators were asked to label the 39 sentences as neutral, negative, or positive. Physicians were given the direction "If you're the physician who wrote this sentence: what is your attitude towards the patient?". Whereas non-physicians were given the direction: "If you're the patient: how do you feel reading this description of you?"

*Unified Labels*
For each sentence, separately for physicians and non-physicians, the sentiment with the most agreement among the annotators was assigned as the unified label. In one instance, the 10 non-physicians were evenly split between neutral and negative labels for the sentence: "However this morning he is adamant that he wants to go to XXX, does not want to go to program and does not want to go home." The lead author decided to label the sentence as neutral.

*Sentiment Analysis Datasets*
To investigate how the language models adapt to the physician/non-physician point of views at different levels of agreement between the two groups, five subsets of labeled sentences were created and split into train and validation with 70/30% ratio for the classification approach and 30/70% ratio for the prompt-based approach:
(1) *Baseline:* All sentences (n=39).
(2) *≥70% agreement*: Sentences with ≥70% agreement within physician/non-physician labels (n=33).
(3) *≥80% agreement*: Sentences with ≥80% agreement within physician/non-physician labels (n=23).
(4) *≥90% agreement*: Sentences with ≥90% agreement within physician/non-physician labels (n=14).
(5) *No agreement:* Sentences with no agreement between physician/non-physician labels (n=8).

It is worth noting, that the subset of sentences with 70% agreement refers to sentences for which at least 7 out of 10 physicians or non-physicians assigned the same label. In other words, a sentence with 70% agreement from non-physicians but 60% agreement from physicians would still be included in the 70% agreement dataset.

The test set for this task consisted of 15 newly selected sentences from the dataset of 1,000 random psychiatric clinical notes that were not already used in this experiment. Each sentence contained at least one word from the patient descriptor lexicon.

*Task Adaptation Dataset*
Sentences in
**Table 1** were removed from the dataset of 1,000 random psychiatric clinical notes. The remaining sentences were then separated into two groups: group one contained words from
**Table 1** (n=368); group two did not contain words from

**Table 1** (n=7,338). The final data set consisted of all sentences in group one, and a subset of sentences from group two (n=65) so that the final dataset had a of 85/15% split between sentences with and without words from Table 1. These sentences were then de-identified and used for task adaptation (TA)[28], where a language model's pre-training corpus is augmented by data from a specific domain not included in the original corpus.

### PLMs with Classification Approach

We utilized three PLMs: RoBERTa[31,32], GatorTron[30], and GatorTron-TA. We task adapted GatorTron with a subset of sentences from psychiatric clinical notes to explore how that impacted model performance. The classification task was designed to ask the models to classify the corresponding sentiment of the sentences from the physician and non-physician point of view. The seed of the models was set to 42.

*Task Adaptation*

Task adaptation (TA), or task-adaptive pretraining, further updates a model's parameters using unlabeled text from a given task[26]. This improves model performance within unique domains, such as psychiatric medical text. In this scenario, we task adapted GatorTron on sentences written about psychiatric patients that were unlabeled and not used throughout the rest of the project (n = 433; Epoch = 5; Batch size = 4), resulting in the GatorTron-TA model.

*Fine-Tuning*

RoBERTa, GatorTron, and GatorTron-TA were fine-tuned to classify the sentiment of sentences labeled by physicians and non-physicians in the Baseline, ≥70% Agreement, and ≥80% Agreement datasets. The ≥90% Agreement dataset was not used due to its low sample size. Hyperparameter optimization was performed during fine-tuning of each model using a grid search approach. Key hyperparameters included learning rate (5e-6 to 1e-4), batch size (2 to 6), number of epochs (1 to 4), weight decay (0 to 0.1), and warm up ratio (0 to 0.1). Each configuration was evaluated on the validation set using macro F1 score, with early stopping applied to prevent overfitting. See **Table 2** for best hyperparameter settings for each model and task. The model fine-tuned with the best settings across each agreement dataset for physician or non-physician labels were saved and evaluated on the same test set (n = 15), wherein the results were manually validated by the psychiatrist (LAL) and computational scientist (AAV) that created the psychiatric patient descriptor lexicon.

### LLMs with Prompt-Based Approach

We utilized three LLMs: GPT-3.5-turbo, Llama-3.1-8B-Instruct, and Mistral-7B-Instruct-v0.2. Prompts were designed to ask the model to classify the corresponding sentiment of the sentences from the physician and non-physician point of view. The temperature of the models was set to 0 or 0.001 and seed to 42.

*Prompt Engineering*

Our ICL prompt-based approach utilized two prompts, one for the physician task and one for the non-physician task, for each model. The prompts can be found in the GitHub folder referenced in Data and Code Availability. In each prompt, we utilized the subsets of training sentences as contextual examples for the model. To determine which training sentences led to the best model's performance, we engineered our prompt using different combinations of negative, neutral, and positive sentences. For example, the training sentences with 80% agreement contained 2 negative, 3 neutral, and 2 positive sentences, so we created a matrix to loop through every combination:

```
[negative:[0,1,2];neutral: [0,1,2,3]; positive:[0,1,2]].
```

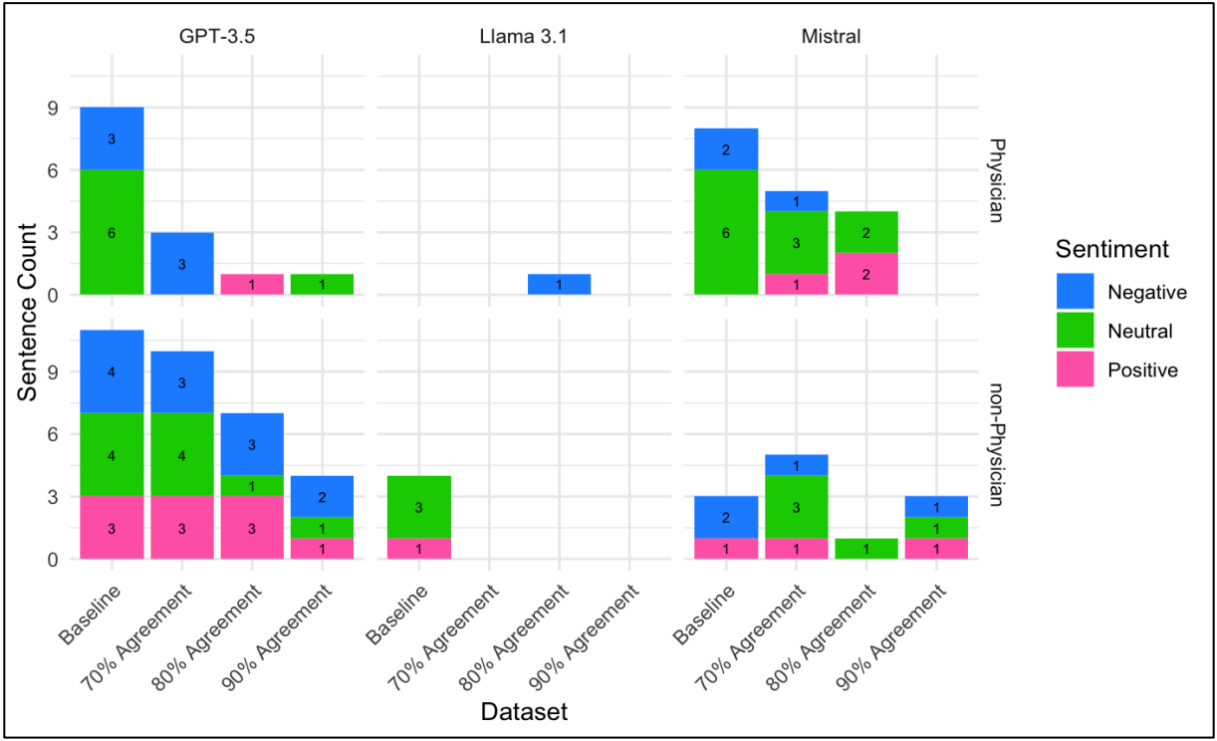

**Figure 4.** Sentence count per sentiment label, model, and dataset. This figure describes the number of sentences across each sentiment label (negative, neutral, positive) included in the prompt for each dataset (Baseline, 70% Agreement, 80% Agreement, 90% Agreement) and LLM that led to the best performance on the validation data for each dataset.

Due to the small number of examples in the subset of sentences with no agreement between physician and non-physician labels, we applied a zero-shot prompt-based approach to that subset. The composition of the best ICL sentences from the training datasets can be found in **Figure 4** and **Table 4**. The ICL sentences that led to the best model performance on the validation task of baseline and 70/80/90% agreement subsets were saved and used with the corresponding prompt for evaluating the model on the same test set as the PLMs. Results were also manually validated by the psychiatrist and computational scientist that created the psychiatric patient descriptor lexicon.

**Ethics Statement**
All sentences extracted from clinical notes underwent manual removal of PHI. The use of the notes was approved by The Mount Sinai Hospital System IRB via STUDY-20-00338: MSCIC Umbrella Protocol. The GPT-3.5 model used is a HIPAA compliant version accessed through the hospital system's API. Llama-3.1 and Mistral were accessed through Hugging Face.

**Author Contributions**
AAV and IL conceived and designed the study. AAV collected the data, performed modeling and manual validation, completed statistics, and edited and wrote the manuscript. LAL performed manual validation, aided in clinical relevance, and edited the manuscript. LC provided feedback on the results and the manuscript. AWC supported the research. IL supervised the research and substantially edited the manuscript. All authors read and approved the final manuscript.

**Acknowledgements**

We thank Ipek Ensari PhD, Ashwin Sawant MD PhD, and Matthew O'Connell PhD for their invaluable contributions to this research as members of the first author's advisory committee. We also thank the annotators who made this work possible. The physician annotators include Alexander Charney, Ali Soroush, Ashwin Sawant, Caroline Massarelli, Donald Apakama, Emma Holmes, Ethan Abbott, Girish Nadkarni, Jihan Ryu, and Lili Chan. The non-physician annotators include Alisha Aristel, Brian Fennessy, Darielle Lewis-Sanders, Eric Vornholt, Eugenia Alessandra Enrica Alleva Bonomi, Maria Koromina, Renata Gonzalez Chong, Rozalyn Wood, Simon Lee, and Tom Kaszemacher. Special appreciation goes to the patients of Mount Sinai – may their data always be used for their benefit.

This work was supported in part through the computational and data resources and staff expertise provided by Scientific Computing and Data at the Icahn School of Medicine at Mount Sinai and supported by the Clinical and Translational Science Awards (CTSA) grant UL1TR004419 from the National Center for Advancing Translational Sciences. Research reported in this publication was also supported by the Office of Research Infrastructure of the National Institutes of Health under award number S10OD026880 and S10OD030463. The content is solely the responsibility of the authors and does not necessarily represent the official views of the National Institutes of Health.

We also acknowledge the funding and resources provided by the Mount Sinai Hospital System and Institute for Personalized Medicine. This study was funded by the US National Institutes of Health grants R01MH121923, R01-PAR-18-896, and IMPACT-MH U01. The funder played no role in study design, analysis and interpretation of data, or the writing of this manuscript.

**Competing Interests**
All authors declare no financial or non-financial competing interests.

**Data Availability**
*The psychiatric patient descriptor lexicon used in this study is available in*
**Table 1** and at: https://github.com/valentinealissa/sentiment_POV/blob/main/lexicon.
The annotated dataset used and analyzed during the current study are available at: https://github.com/valentinealissa/sentiment_POV/blob/main/sentiment_sentences_annotated.csv.

**Code Availability**
The underlying code and prompts used for this study is available in "sentiment_POV" and can be accessed via this link: https://github.com/valentinealissa/sentiment_POV.

**Supplementary Material**

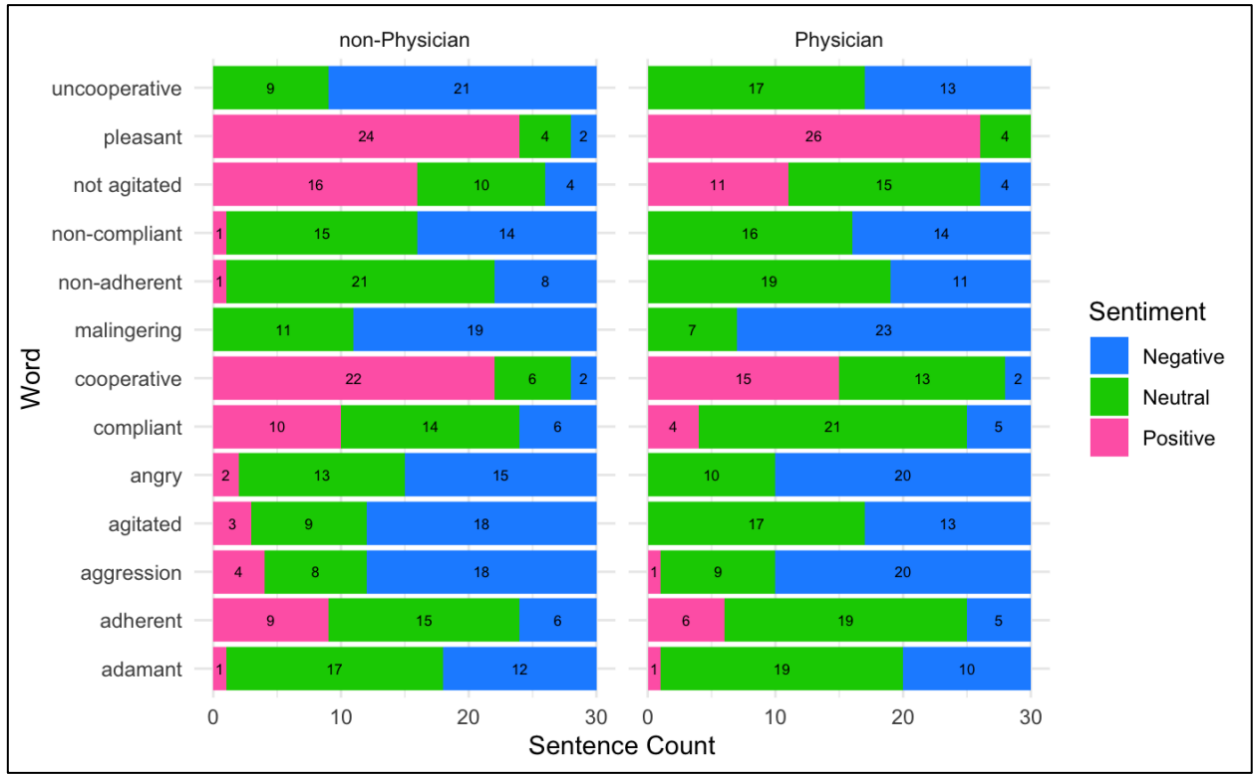

**Figure 5.** Sentence count per sentiment label, word from the patient descriptor lexicon, and rater group (physician or non-physician). This figure describes the number of sentences containing each word in the patient descriptor lexicon that obtained a negative, neutral, or positive label from the unified physician or non-physician labels.

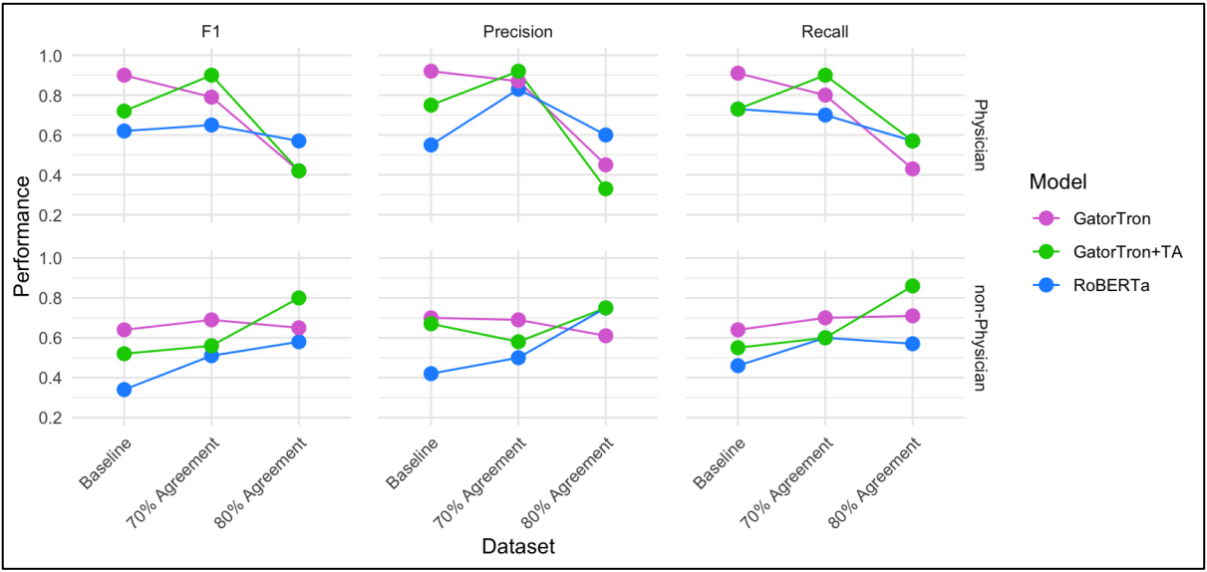

**Figure 6.** Pretrained language models (PLMs) performance on validation set. The performance of each PLM, i.e., GatorTron, GatorTron with task adaptation (GatorTron+TA), and RoBERTa is evaluated on the validation set after fine-tuning on the training dataset using the best hyperparameters. Macro F1 score, precision, and recall are reported.

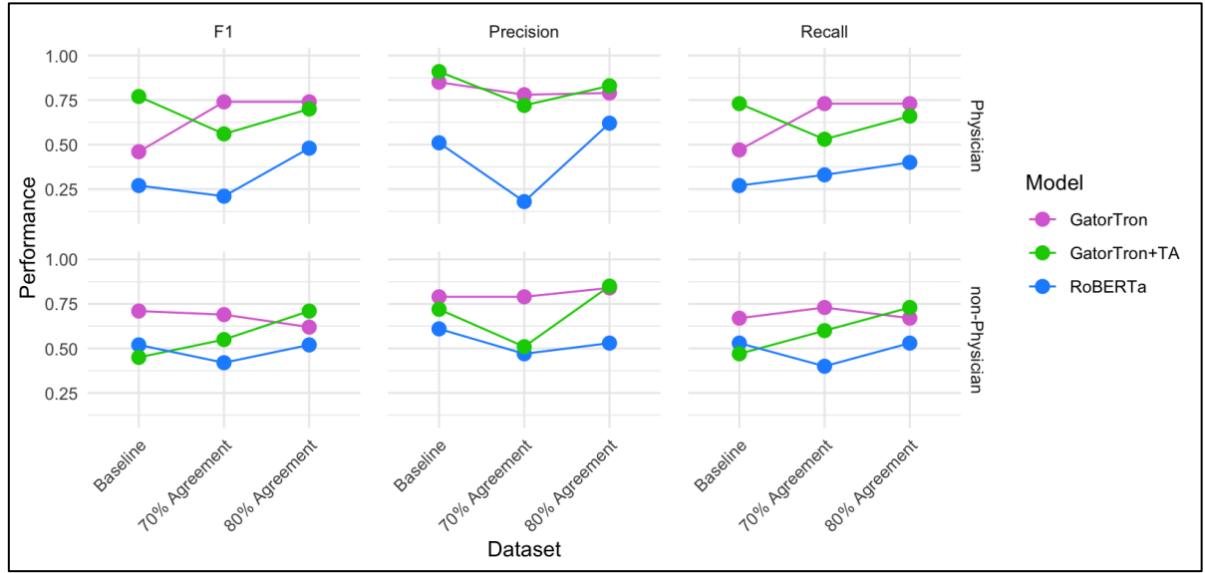

**Figure 7.** Pretrained language models performance on test set. Each graph demonstrates the performance of each model, i.e., GatorTron, GatorTron with task adaptation (GatorTron+TA), and RoBERTa on the test set after fine-tuning. Macro F1 score, precision, and recall are reported.

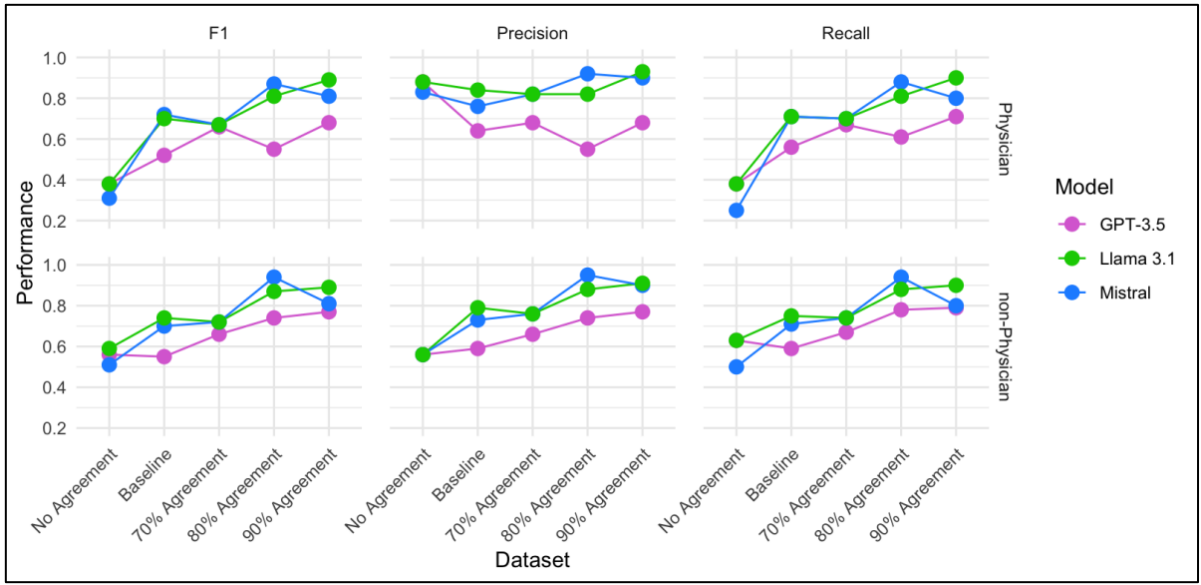

**Figure 8.** Large language model (LLM) performance on validation set with zero-shot approach. Each graph shows the F1 score, precision, and recall for each LLM (i.e., GPT-3.5, Llama-3.1, and Mistral) on the validation data for each dataset (i.e., No Agreement, Baseline, 70% Agreement, 80% Agreement, 90% Agreement) using a zero-shot approach. Macro F1 score, precision, and recall are reported.

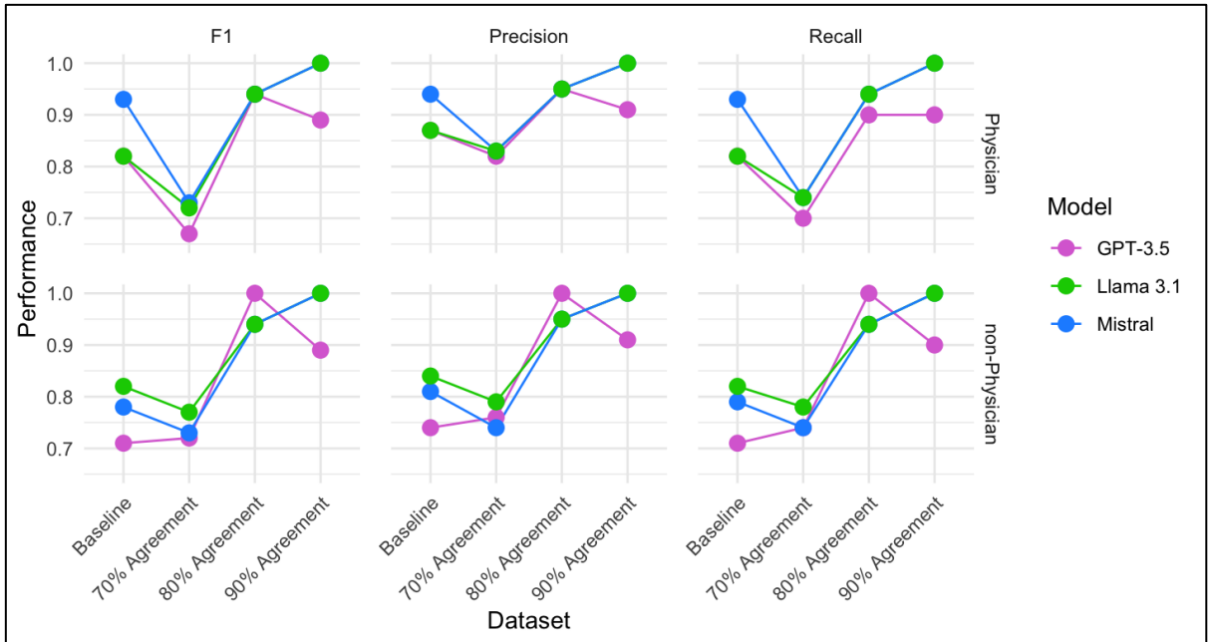

**Figure 9.** Large language model (LLM) performance on validation set with in-context learning (ICL) approach. Each graph shows the macro F1 score, precision, and recall for each LLM (i.e., GPT-3.5, Llama-3.1, and Mistral) on the validation data for each dataset (i.e., No Agreement, Baseline, 70% Agreement, 80% Agreement, 90% Agreement) using a prompt with the best ICL sentences from the corresponding training dataset.

| Word | Sentence |
|---|---|
| Adamant | Patient very adamantly against hospitalization; states she is not suicidal and needs to go home to care for her dog |
| Adherent | States he is adherent w/ his outpatient XXX and HIV meds. |
| Agitated | She has been awake now for several hours and has become active and hyperverbal, not overly agitated but mood labile |
| Aggressive | Pt remains aggressive and very threatening upon arrival, tried to hit security officer with his head while still in handcuff |
| Angry | He remained irritable with an angry edge but was able to respond to redirections |
| Compliant | She states that pt has been compliant with meds |
| Cooperative | Calm and cooperative, agrees with plan to stay overnight and went back to sleep |
| Malingering | Pt had been evaluated in XXX ED earlier today, and was felt to be malingering re: XXX complaints leading her to request evaluation/admission |
| Non-Adherent | Given recent non-adherence, will restart pt on VPA 500mg BID, fluphenazine 5mg PO BID and benztropine 1mg BID |
| Not Agitated | On reassessment this AM, pt remained calm, not agitated, and again is without SI/HI/AH/VH/PI or delusional content |
| Non-Compliant | His worsening psychotic symptom secondary to medication non compliance and substance use (utox (+) cocaine/cannabis), will admit for safety |
| Pleasant | MSE: pleasant, cooperative, euthymic, speech wnl, affect full and appropriate to content |
| Uncooperative | He is not cooperative with questions and starts screaming incoherently "which is it, which is it, which is it" unable to re-direct after this |

**Table 1.** Lexicon and sample sentences.

| Model | POV | Dataset | Learning Rate | Batch Size | Epochs | Weight Decay | Warm Up Ratio | F1 | Precision | Recall |
|---|---|---|---|---|---|---|---|---|---|---|
| RoBERTa | Physician | Baseline | 5e-5 | 4 | 1 | 0.1 | 0 | 0.62 (0.27) | 0.55 (0.51) | 0.73 (0.27) |
| | | 70% | 5e-5 | 4 | 2 | 0 | 0 | 0.65 (0.21) | 0.85 (0.18) | 0.70 (0.33) |
| | | 80% | 5e-6 | 4 | 1 | 0 | 0 | 0.57 (0.48) | 0.60 (0.62) | 0.57 (0.40) |
| | Non-Physician | Baseline | 1e-4 | 4 | 1 | 0 | 0 | 0.34 (0.52) | 0.42 (0.61) | 0.46 (0.53) |
| | | 70% | 1e-4 | 4 | 1 | 0 | 0.1 | 0.51 (0.42) | 0.50 (0.47) | 0.60 (0.40) |
| | | 80% | 5e-5 | 4 | 1 | 0.1 | 0 | 0.58 (0.52) | 0.75 (0.53) | 0.57 (0.53) |
| GatorTron | Physician | Baseline | 5e-5 | 4 | 1 | 0 | 0.1 | 0.90 (0.46) | 0.92 (0.85) | 0.91 (0.47) |
| | | 70% | 5e-5 | 4 | 2 | 0 | 0 | 0.80 (0.74) | 0.87 (0.78) | 0.80 (0.73) |
| | | 80% | 5e-5 | 4 | 1 | 0.1 | 0.1 | 0.43 (0.74) | 0.45 (0.79) | 0.43 (0.73) |
| | Non-Physician | Baseline | 2e-5 | 4 | 4 | 0 | 0 | 0.64 (0.71) | 0.70 (0.79) | 0.64 (0.67) |
| | | 70% | 1e-4 | 4 | 1 | 0.1 | 0.1 | 0.69 (0.69) | 0.69 (0.79) | 0.70 (0.73) |
| | | 80% | 1e-4 | 4 | 4 | 0 | 0 | 0.65 (0.62) | 0.61 (0.84) | 0.71 (0.67) |
| GatorTron-TA | Physician | Baseline | 5e-5 | 4 | 1 | 0 | 0.1 | 0.90 (0.77) | 0.92 (0.91) | 0.90 (0.73) |
| | | 70% | 1e-4 | 4 | 2 | 0 | 0 | 0.90 (0.56) | 0.92 (0.72) | 0.90 (0.53) |
| | | 80% | 5e-6 | 4 | 1 | 0 | 0 | 0.42 (0.70) | 0.33 (0.83) | 0.57 (0.66) |
| | Non-Physician | Baseline | 5e-5 | 4 | 4 | 0.1 | 0 | 0.54 (0.45) | 0.67 (0.72) | 0.55 (0.47) |
| | | 70% | 1e-4 | 4 | 1 | 0.1 | 0 | 0.56 (0.55) | 0.58 (0.51) | 0.60 (0.60) |
| | | 80% | 5e-6 | 4 | 1 | 0 | 0 | 0.80 (0.71) | 0.75 (0.85) | 0.86 (0.73) |

**Table 3.** Pretrained language models performance during validation (test performance in parentheses) with the best hyperparameters.

| | Physician Raters | non-Physician Raters |
|---|---|---|
| **Negative** | 75% | 79% |
| **Neutral** | 69% | 63% |
| **Positive** | 80% | 77% |

**Table 2.** Average percent agreement of unified label per sentence for sentiment label and rater group.

| Model | POV | Dataset | Negative Sentences | Neutral Sentences | Positive Sentences | F1 | Precision | Recall |
|---|---|---|---|---|---|---|---|---|
| GPT-3.5 | Physician | Baseline | 3 | 6 | 0 | 0.82 (0.84) | 0.87 (0.90) | 0.82 (0.87) |
| | | 70% | 3 | 0 | 0 | 0.67 (0.79) | 0.82 (0.79) | 0.70 (0.80) |
| | | 80% | 0 | 0 | 1 | 0.94 (0.93) | 0.95 (0.94) | 0.90 (0.93) |
| | | 90% | 0 | 1 | 0 | 0.89 (0.84) | 0.91 (0.90) | 0.90 (0.87) |
| | Non-Physician | Baseline | 4 | 4 | 3 | 0.71 (0.75) | 0.74 (0.85) | 0.71 (0.80) |
| | | 70% | 3 | 4 | 3 | 0.72 (0.75) | 0.76 (0.85) | 0.74 (0.80) |
| | | 80% | 3 | 1 | 3 | 1.00 (0.85) | 1.00 (0.89) | 1.00 (0.87) |
| | | 90% | 2 | 1 | 1 | 0.89 (0.85) | 0.91 (0.89) | 0.90 (0.87) |
| Llama 3.1 | Physician | Baseline | 0 | 0 | 0 | 0.82 (0.71) | 0.87 (0.64) | 0.82 (0.80) |
| | | 70% | 0 | 0 | 0 | 0.72 (0.71) | 0.83 (0.64) | 0.74 (0.80) |
| | | 80% | 1 | 0 | 0 | 0.94 (0.71) | 0.95 (0.64) | 0.94 (0.80) |
| | | 90% | 0 | 0 | 0 | 1.00 (0.71) | 1.00 (0.64) | 1.00 (0.80) |
| | Non-Physician | Baseline | 0 | 3 | 1 | 0.82 (0.79) | 0.84 (0.79) | 0.82 (0.80) |
| | | 70% | 0 | 0 | 0 | 0.77 (0.62) | 0.79 (0.54) | 0.78 (0.73) |
| | | 80% | 0 | 0 | 0 | 0.94 (0.62) | 0.95 (0.54) | 0.94 (0.73) |
| | | 90% | 0 | 0 | 0 | 1.00 (0.62) | 1.00 (0.54) | 1.00 (0.73) |
| Mistral | Physician | Baseline | 2 | 6 | 0 | 0.93 (0.76) | 0.94 (0.84) | 0.93 (0.73) |
| | | 70% | 1 | 3 | 1 | 0.73 (0.81) | 0.83 (0.86) | 0.74 (0.80) |
| | | 80% | 0 | 2 | 2 | 0.94 (0.76) | 0.95 (0.84) | 0.94 (0.73) |
| | | 90% | 0 | 0 | 0 | 1.00 (0.76) | 1.00 (0.84) | 1.00 (0.73) |
| | Non-Physician | Baseline | 2 | 0 | 1 | 0.78 (0.93) | 0.81 (0.94) | 0.79 (0.93) |
| | | 70% | 1 | 3 | 1 | 0.73 (0.87) | 0.74 (0.89) | 0.74 (0.87) |
| | | 80% | 0 | 1 | 0 | 0.94 (0.87) | 0.95 (0.91) | 0.94 (0.87) |
| | | 90% | 1 | 1 | 1 | 1.00 (0.87) | 1.00 (0.89) | 1.00 (0.87) |

**Table 4.** Large language models performance during validation (test performance in parentheses) with the best sentences used in in-context learning (ICL).